\documentclass{article}

\usepackage{arxiv}

\usepackage[utf8]{inputenc} 
\usepackage[T1]{fontenc}    
\usepackage{hyperref}       
\usepackage{url}            
\usepackage{booktabs}       
\usepackage{amsfonts}       
\usepackage{nicefrac}       
\usepackage{microtype}      
\usepackage{lipsum}		
\usepackage{graphicx}
\usepackage{natbib}
\usepackage{doi}

\usepackage{subfig}
\usepackage{amsmath}
\usepackage{algorithm}
\usepackage{algorithmic}

\title{Artificial Intelligence-Driven Clinical Decision Support Systems}

\author{ {Muhammet Alkan} \\
	School of Computing Science\\
	University of Glasgow\\
	Glasgow, Scotland, UK \\
    \And
	{Idris Zakariyya} \\
	School of Computing Science\\
	University of Glasgow\\
	Glasgow, Scotland, UK \\
    \And
	{Samuel Leighton} \\
	School of Health and Well Being\\
	University of Glasgow\\
	Glasgow, Scotland, UK \\
    \And
	{Kaushik Bhargav Sivangi} \\
	School of Computing Science\\
	University of Glasgow\\
	Glasgow, Scotland, UK \\
    \And
	{Christos Anagnostopoulos} \\
	School of Computing Science\\
	University of Glasgow\\
	Glasgow, Scotland, UK \\
    \And
	{Fani Deligianni}\thanks{Corresponding author, \texttt{Fani Deligianni}} \\
	School of Computing Science\\
	University of Glasgow\\
	Glasgow, Scotland, UK \\
    \texttt{Fani.Deligianni@glasgow.ac.uk}
}

\renewcommand{\shorttitle}{Artificial Intelligence-Driven Clinical Decision Support Systems}

\begin{document}
\maketitle

\begin{quote}
    ``The advance of technology is based on making it fit in so that you don't really even notice it, so it's part of everyday life.'' --- Bill Gates
\end{quote}

\begin{abstract}
As artificial intelligence (AI) becomes increasingly embedded in healthcare delivery, this chapter explores the critical aspects of developing reliable and ethical Clinical Decision Support Systems (CDSS). Beginning with the fundamental transition from traditional statistical models to sophisticated machine learning approaches, this work examines rigorous validation strategies and performance assessment methods, including the crucial role of model calibration and decision curve analysis. The chapter emphasizes that creating trustworthy AI systems in healthcare requires more than just technical accuracy; it demands careful consideration of fairness, explainability, and privacy.
The challenge of ensuring equitable healthcare delivery through AI is stressed, discussing methods to identify and mitigate bias in clinical predictive models. The chapter then delves into explainability as a cornerstone of human-centered CDSS. This focus reflects the understanding that healthcare professionals must not only trust AI recommendations but also comprehend their underlying reasoning.
The discussion advances in an analysis of privacy vulnerabilities in medical AI systems, from data leakage in deep learning models to sophisticated attacks against model explanations. The text explores privacy-preservation strategies such as differential privacy and federated learning, while acknowledging the inherent trade-offs between privacy protection and model performance. 
This progression, from technical validation to ethical considerations, reflects the multifaceted challenges of developing AI systems that can be seamlessly and reliably integrated into daily clinical practice while maintaining the highest standards of patient care and data protection.

\end{abstract}


\section{Artificial Intelligence-Driven Clinical Decision Support Systems}

\subsection{From machine learning and statistical models to clinical decision support systems: An Overview}

Clinical research demands a meticulous, multifaceted approach when evaluating predictive models, which extends beyond the traditional data science perspective. As we dive into this complex landscape, we must consider a series of key questions that shape the development and validation of machine learning models, ultimately determining their suitability as decision support systems in healthcare. 

In prediction modeling, our main focus is on estimating the risk of adverse events based on a combination of factors. We seek to understand not only the predictive power of these factors, but also their individual contributions to the model's decision-making process. This understanding is crucial, as it allows us to incorporate subject matter knowledge into the modeling pipeline, bridging the gap between data-driven insights and clinical expertise.

The foundations of the clinical prediction model validation process have been presented in Steyerberg \citep{steyerberg2014towards}.
At the heart of this process lies the fundamental research question or hypothesis. For example, the choice of prediction outcome is paramount in clinical research. 
Outcomes such as mortality rates at 30 days are frequently relevant to various research questions. However, it's not just the nature of the outcome that matters, but also its frequency within the dataset. This frequency effectively determines the sample size, which in turn influences the statistical power and reliability of the model.

The selection of patient data for the development of the model is a critical consideration. Often, these data are collected for purposes other than the study at hand, raising questions about their representativeness. We must carefully examine whether patient records truly reflect the population for which the study is intended. Additionally, the treatment of prognostic factors and their effects presents a unique challenge. Although traditional studies often consider treatment effects negligible compared to prognostic factors, there are instances where these effects warrant specific attention. Adjusting for baseline prognostic factors can offer significant advantages in estimating treatment effects applicable to individual patients.

The reliability and completeness of the predictor measurements pose another hurdle in model development. Incomplete datasets are common, with missing values for potential predictors. The approach to handling these missing data can significantly impact the model's performance and validity. Although complete case analysis – excluding patients with missing values – is a straightforward solution, it often results in the loss of significant information. More sophisticated methods, such as imputation techniques that leverage correlations between variables, offer a more nuanced approach to preserving data integrity. In addition, informative missingness, where the fact that data are missing is related to the outcome of interest, must be carefully considered to avoid biased results.

As we navigate these considerations, we recognize that the development of machine learning models for clinical decision support systems is a multifaceted process. It requires a delicate balance between statistical rigor, clinical relevance, and practical applicability. By addressing these key questions and challenges, we pave the way for more robust and reliable predictive models that can enhance clinical decision-making and, ultimately, patient care.

\subsection{A Quick Overview of Model Development and Validation Strategies of Machine Learning Models}

\begin{figure}[h]
    \centering
    \includegraphics[width=0.8\linewidth]{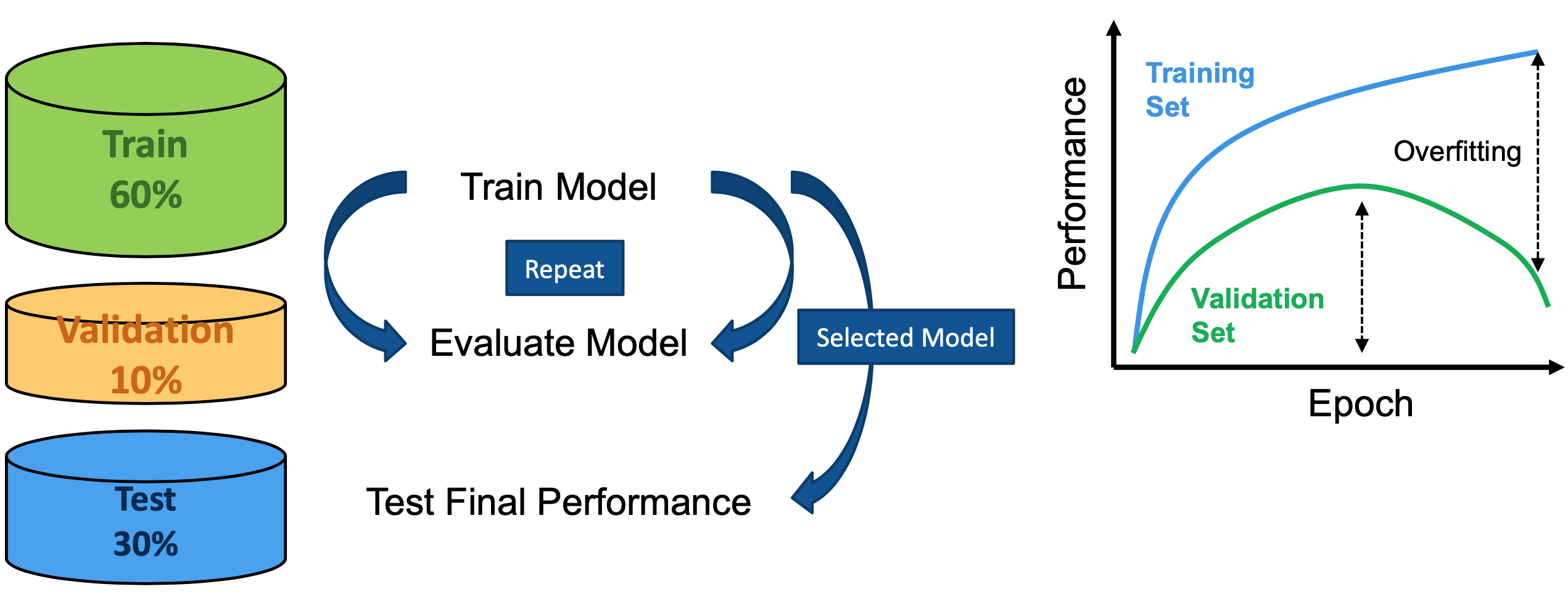}
    \caption{Model Development}
    \label{fig:ModelDevelopment}
\end{figure}

Model evaluation and selection are critical steps in the machine learning development process. In an ideal scenario, we would have access to data that perfectly represents the entire target population. In this case, we could train and test the machine learning model using the same data, and the error rate obtained would closely reflect the true error rate when the number of samples is very large.
However, in reality, the error rate obtained when training and testing on the same dataset is positively biased. This is because the model has been exposed to the same data during both the training and testing phases, which can lead to an overly optimistic estimate of its performance.
To address this issue, in real-world applications, it is important to split the available data into two separate sets: a training set and a testing set. Typically, around 70\% of the data is used for training the model, while the remaining 30\% is reserved for testing its performance on unseen examples. 
By separating the training and testing data, we can evaluate the model's ability to make accurate predictions on new, unseen data, which is crucial for deploying the model in real-world applications. 

In most cases, we estimate the empirical risk based on a limited number of samples. This involves measuring the loss function with respect to our trained classifier, as discussed in \citep{japkowicz2011evaluating}. Empirical risk estimation is achieved by computing the average loss over the data points ($m$ is the number of samples) according to a loss function $L$, which penalizes the differences between the predicted values $f(x)$ and the actual targets $y$:

\begin{equation}
\label{eq:EmpiricalRisk}
R_S(f)={\frac {1}{m}}\sum _{i=1}^{m}L(y_{i}, f(x_{i}))
\end{equation}

Variations in the empirical risk estimation can arise from several factors. These include random variations in the training and testing sets, the learning algorithm itself, and even the noise inherent in the data classes being considered. 
One key advantage of the hold-out method (where the training and testing sets are independent) is that it provides some guarantees about the model's performance on data it has not been trained on before. However, we must also consider the confidence intervals around the empirical risk estimation. For example, when evaluating a machine learning algorithm, we should not assume a Gaussian distribution of the loss error, as the errors may be clustered near zero.

The error can be modeled using the Bernoulli distribution to calculate the error bound, Equation \ref{eq:ConfidenceInterval}, providing an indication of the potential deviation between the empirical risk estimation and the true risk with a probability of accuracy of ($1-\delta$). In Equation \ref{eq:ConfidenceInterval}, $m'$ represents the sample size and $\delta$ represents the confidence parameter, which quantifies the probability that our estimate is accurate.

\begin{equation}
\label{eq:ConfidenceInterval}
E = \sqrt{\frac{1}{2m^{'}}\ln(\frac{2}{\delta})}
\end{equation}

\begin{figure}[h]
    \centering
    \includegraphics[width=1.0\linewidth]{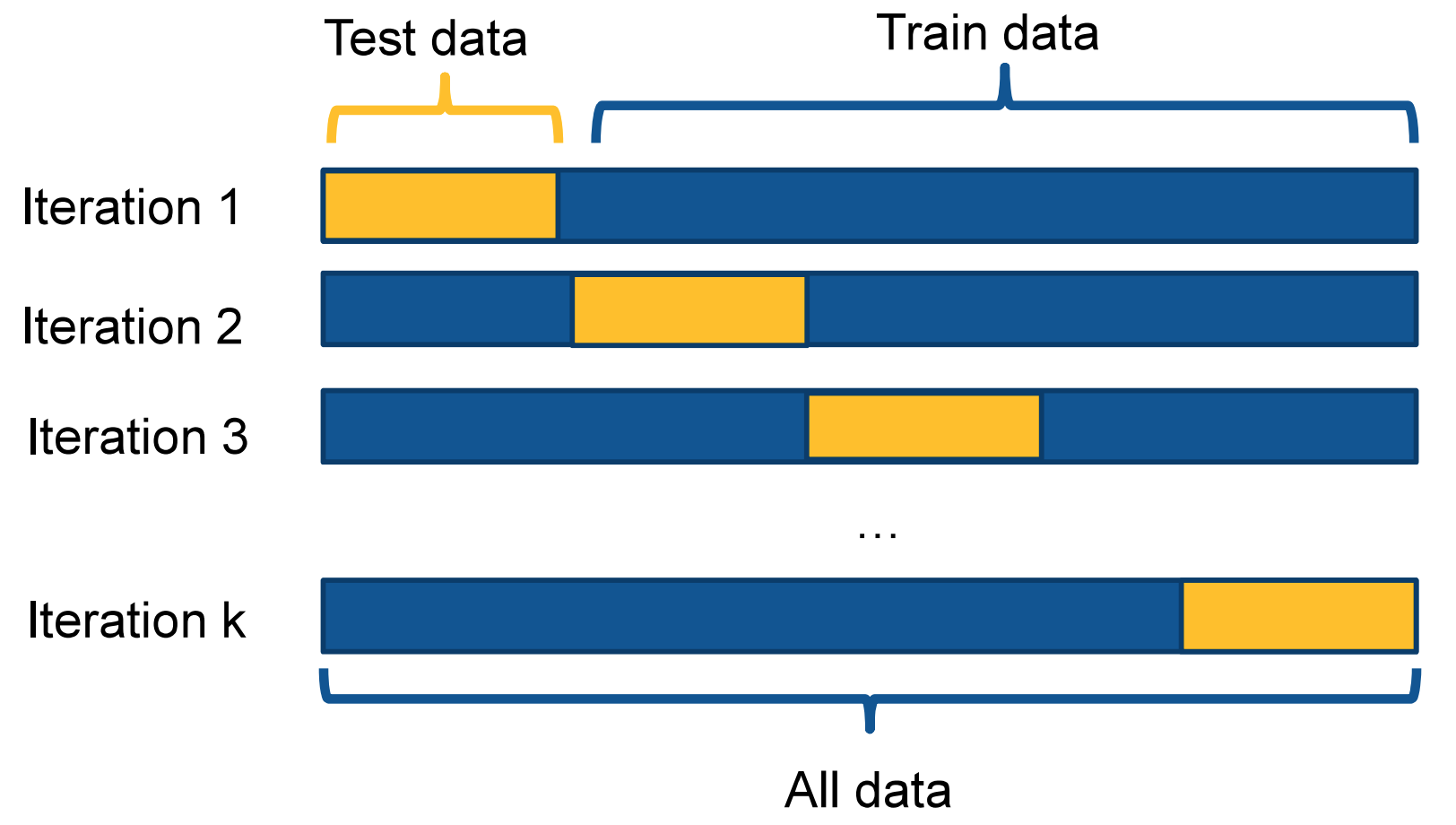}
    \caption{$k$-fold cross validation}
    \label{fig:kFold}
\end{figure}

The $k$-fold cross-validation is one of the most popular error estimation approaches in machine learning model training and evaluation.
In this method, the dataset is divided into $k$ distinct parts or "folds" like in Figure \ref{fig:kFold}. During each iteration, one of these folds is reserved for testing, while the remaining ($k-1$) folds are used for training the model.
This process is repeated $k$ times, with a different fold serving as the test set each time. By doing so, we obtain $k$ separate estimates of the classifier's error rate. These estimates can then be averaged to give the mean performance of the algorithm across the different folds.
Examining the variability of the error estimates across the $k$ iterations can also provide valuable insights into the algorithm's stability and robustness.
A key advantage of $k$-fold cross-validation is that the test samples are independent between the different folds, as there is no overlap. This helps to ensure a more reliable and unbiased assessment of the model's generalization capabilities.

A potential issue that can arise with standard $k$-fold cross-validation is that the data may not be evenly distributed across classes. 
This problem becomes worse when dealing with imbalanced class data, which is common in healthcare applications. To address this, we can employ a technique called stratified $k$- fold cross validation. In this approach, the folds are created in a way that ensures the class distribution within each fold closely matches the original class distribution in the overall dataset.

A special case of $k$-fold cross-validation is leave-one-out cross validation (LOOCV). In this case, the value $k$ is set to the number of samples in the dataset, meaning that each sample is used as the test set once while the remaining ($k-1$) samples are used for training. 
LOOCV has the advantage of utilising most of the available data, which can result in relatively unbiased estimate of the model's performance. However, this comes at the cost of significant computational expense, especially as the dataset size grows. 
Although, LOOCV may provide better performance estimates in datasets with extreme values, it is important to note that this is not a guarantee of an unbiased classifier, especially when dealing with small datasets. The underlying assumption of LOOCV is that the training set is representative of the true data distribution, which may not always hold.

One key aspect of validation techniques such as $k$-fold cross-validation and LOOCV is that the estimate is not based on a single, fixed classifier. Instead, the model is retrained each time, producing a new classifier with each iteration. This approach has both advantages and disadvantages. The primary advantage is that it allows us to assess the stability of machine learning models across different data partitions. However, the disadvantage is that when comparing the performance of different algorithms, we must remember that we are comparing the average performance estimates of various classifiers, rather than a single, fixed classifier as in the holdout method. To address this, nested cross-validation is often used, as it provides a more robust estimate of model performance by incorporating an additional layer of cross-validation to tune hyperparameters, thereby reducing the risk of overfitting.

\begin{figure}[h!]
    \centering
    \subfloat[\centering \label{fig:ConfusionMatrix}Confusion matrix] 
    {{\includegraphics[height=0.28\textwidth]{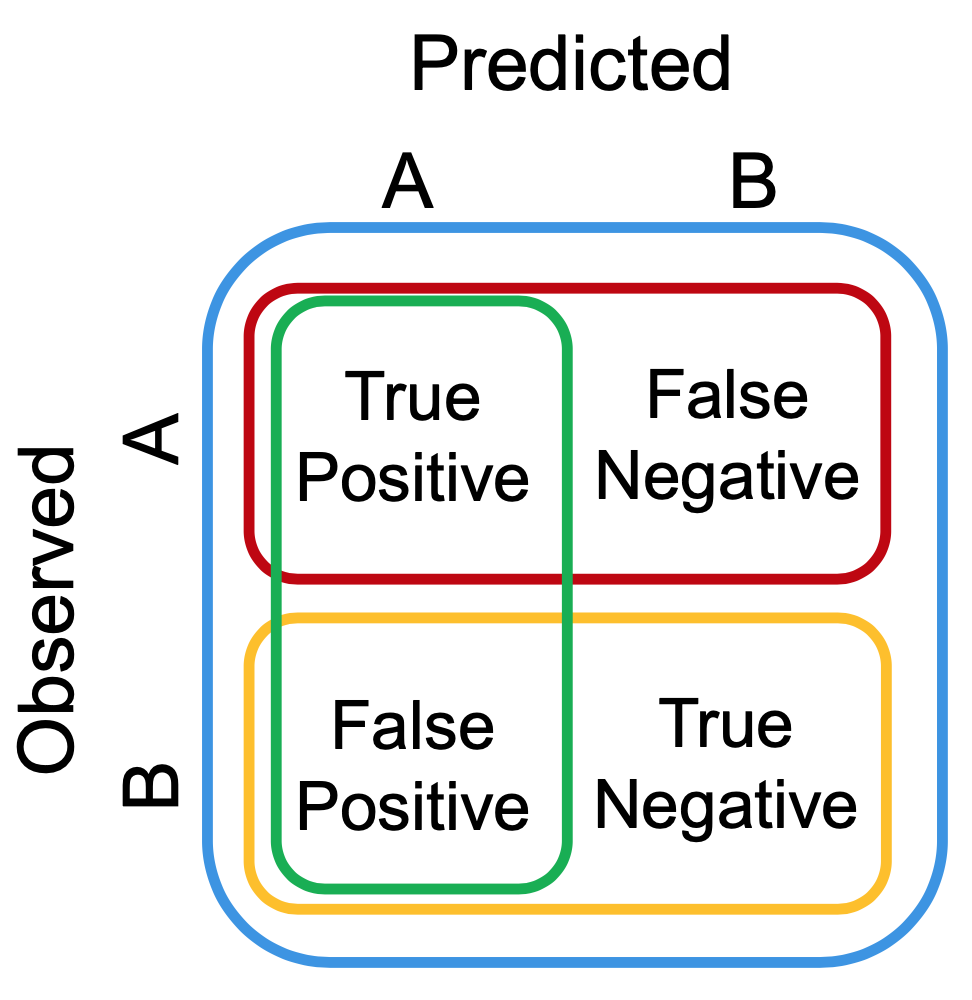} }}
    \hspace{0.1cm}
    \subfloat[\centering \label{fig:PerformanceMetrics}Performance metrics] 
    {{\includegraphics[height=0.28\textwidth]{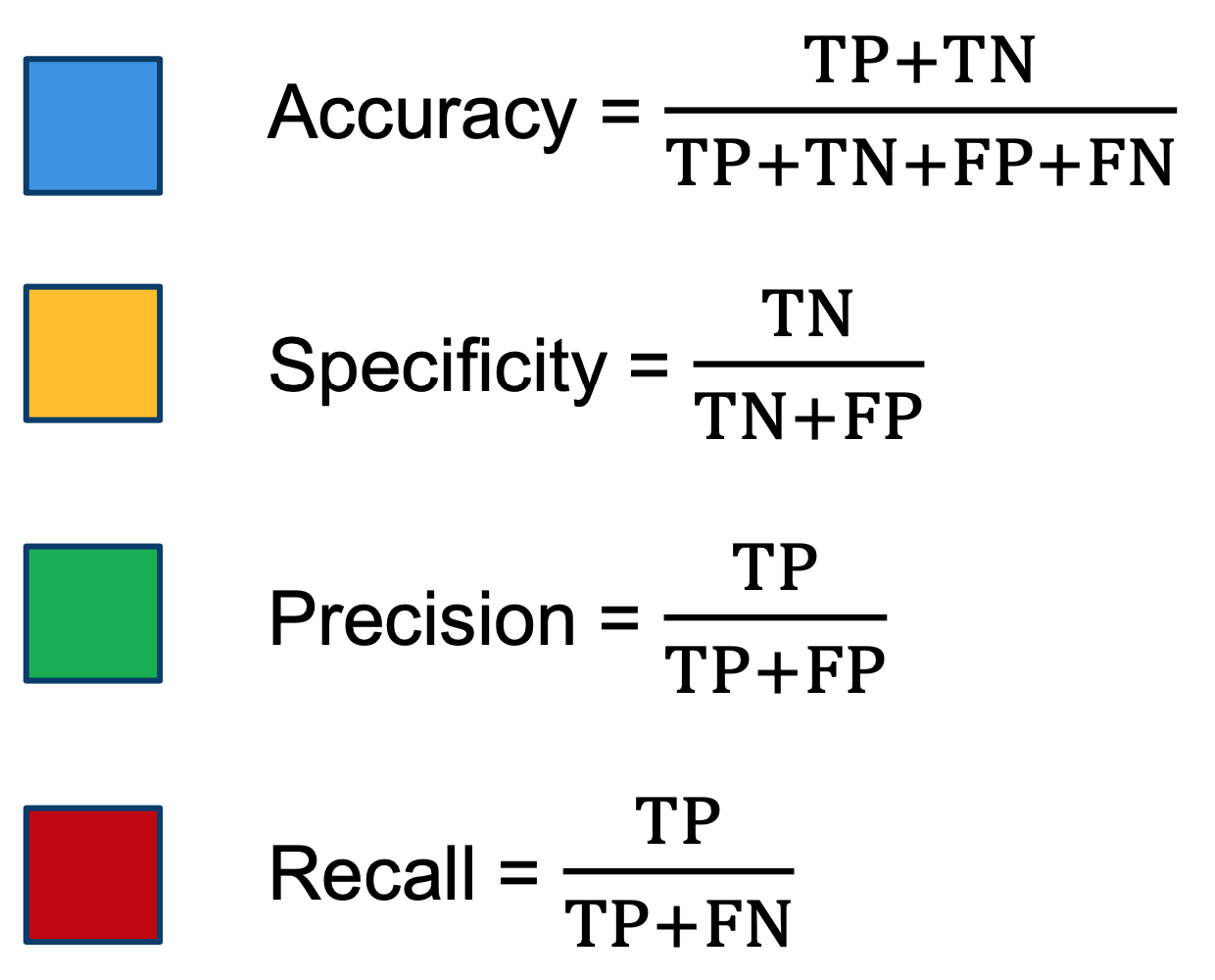} }}
    \hspace{0.1cm}
    \subfloat[\centering \label{fig:ROC_AUC}The ROC curve] 
    {{\includegraphics[height=0.28\textwidth]{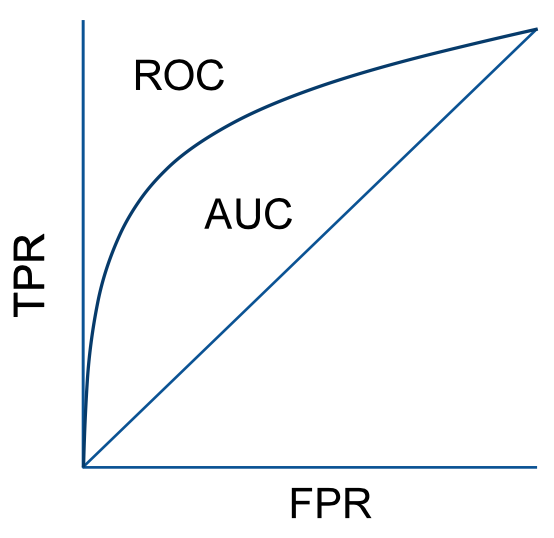} }}
    \caption{ Performance evaluation of ML models.}
    \label{fig:evaluationML}
\end{figure}

One way to describe the performance of classification algorithms is through a confusion matrix as in Figure \ref{fig:PerformanceMetrics}(a). This square matrix has rows and columns equal to the number of classes. The diagonal elements represent the true positives and true negatives, assuming "positive" refers to one class and "negative" to another. The off-diagonal elements indicate false positives and false negatives. 
 From the confusion matrix, we can derive several performance metrics. For instance, accuracy is the ratio of correctly predicted observations to the total observations. Specificity, or the true negative rate, shows how well the classifier identifies negative cases, while recall also called sensitivity, or the true positive rate, indicates how well it identifies positive cases. Precision reflects the positive predictive value for a class. The F1 score combines recall and precision into a single metric, weighting them evenly.

An alternative method to assess the performance of a machine learning algorithm is by using a Receiver Operating Characteristic (ROC) curve, as shown in Figure \ref{fig:ROC_AUC}. 
The ROC curve plots the false positive rate on the horizontal axis and the true positive rate on the vertical axis. The true positive rate, or sensitivity, indicates how well the classifier identifies positive cases, while the false positive rate can be expressed as $(1 - \mathit{specificity})$. Thus, the ROC curve illustrates the trade-off between sensitivity and specificity for the classifier. It is generated by varying the threshold for classifying positive and negative cases. This makes the ROC curve a comprehensive measure of performance, as it considers different threshold settings. It is used not only to analyze the behavior of machine learning algorithms but also for model selection by identifying the optimal threshold region. For a random classifier, the ROC curve would be a straight diagonal line. The area under the curve (AUC) summarizes the classifier's performance, with higher values indicating better performance. The AUC is often used to compare different classifiers. Additionally, there are various extensions of the ROC and AUC for multi-class scenarios.

When comparing the performance of one algorithm against another, or multiple algorithms across one or more datasets, it is common to use null hypothesis statistical testing. Several statistical tests can validate the performance of two algorithms, but it is crucial to consider the assumptions underlying these tests. For instance, the paired t-test assumes a normal distribution, independence of measurements, and an adequate sample size. If the normality assumption is violated, non-parametric tests are often used. One such test is the Wilcoxon signed-rank test, an alternative to the paired t-test. This test is based on the ranks of the absolute differences, making it more robust to outliers. It is important to note that both parametric and non-parametric tests can be manipulated by increasing the number of samples, potentially affecting the results of the null hypothesis statistical testing approach.

\subsection{Performance Validation in Clinical Decision Support Systems}




Validation of prediction models tailored for clinical decision support systems should occur through both internal and external methods \citep{steyerberg2014towards, ramspek2021external}. Internal validation, using techniques like split-sample validation, cross-validation, or bootstrapping, assesses reproducibility within the development population. External validation, involving patients from different populations, tests the model's generalizability across various settings and demographics.

While internal validation techniques provide valuable insights into a model's performance, external validation serves as a crucial complement in assessing predictive models for clinical use. This process involves testing the model on data that is entirely separate from the development dataset, often collected from different institutions or time periods. 
Despite the growing number of publications on prediction models, studies employing both internal and external validation remain relatively scarce. This highlights the challenges in establishing predictive models as reliable decision support systems.

\begin{figure*}
\centering
\includegraphics[width=1\linewidth]{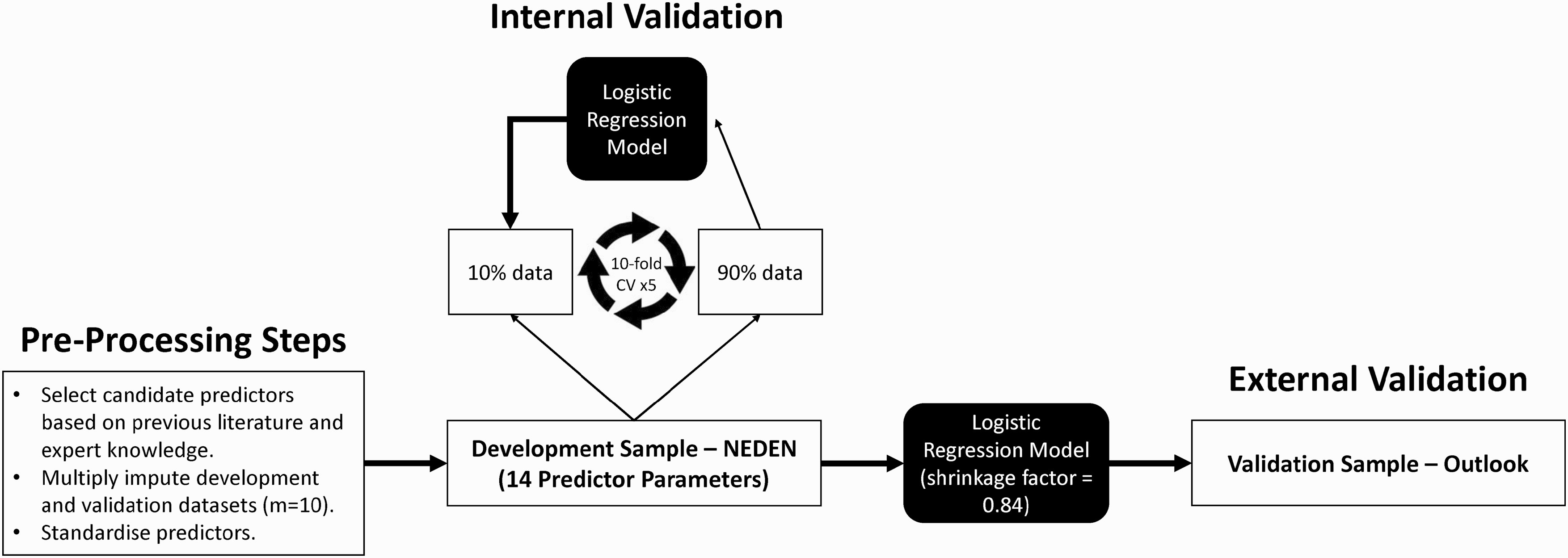}
\caption{\centering A Nonremission Risk Prediction Model in First-Episode Psychosis (reproduced with permission \citep{leighton2021development}) }
\label{fig:ModelPsychosis}
\end{figure*}

Figure \ref{fig:ModelPsychosis} illustrates a risk prediction tool designed to estimate the likelihood of symptom nonremission in first-episode psychosis \citep{leighton2021development}. The tool's internal validation performance was assessed using ten-fold cross-validation yielding an AUC of 0.74. External validation was conducted with patient data from various sites, produced an AUC of 0.73, confirming the model's generalisability.  
Successful external validation strengthens confidence in a model's clinical utility. It demonstrates that the model's predictions remain accurate across different patient populations and healthcare settings. This robustness is essential for establishing the model as a trustworthy component of clinical decision support systems. 
In other words, external validation offers a more rigorous test of a model's generalizability, revealing how well it performs in diverse real-world scenarios. It helps identify potential overfitting issues that may not be apparent through internal validation alone. 
In essence, external validation acts as a bridge between theoretical model development and practical clinical application. It provides the evidence needed to justify the integration of predictive models into healthcare decision-making processes, ultimately contributing to improved patient care and outcomes.

\subsection{Calibration of Clinical Prediction Models}

In clinical practice, the validation of machine learning models extends beyond traditional prediction performance metrics. While measures like AUC, precision, and F1 scores are crucial, they do not fully capture a model's clinical utility. Effective clinical decision support systems require assessment of underlying risk estimates and clinical usefulness, which can be subjective and application-dependent.
Model calibration is a critical aspect of validation, referring to the agreement between observed outcomes and predictions \citep{van2019calibration}. For instance, if a model predicts a 15\% risk of 30-day mortality, approximately 15 out of 100 patients with such a prediction should experience the outcome. 

\begin{figure*}[h]
\centering
\includegraphics[width=0.7\linewidth]{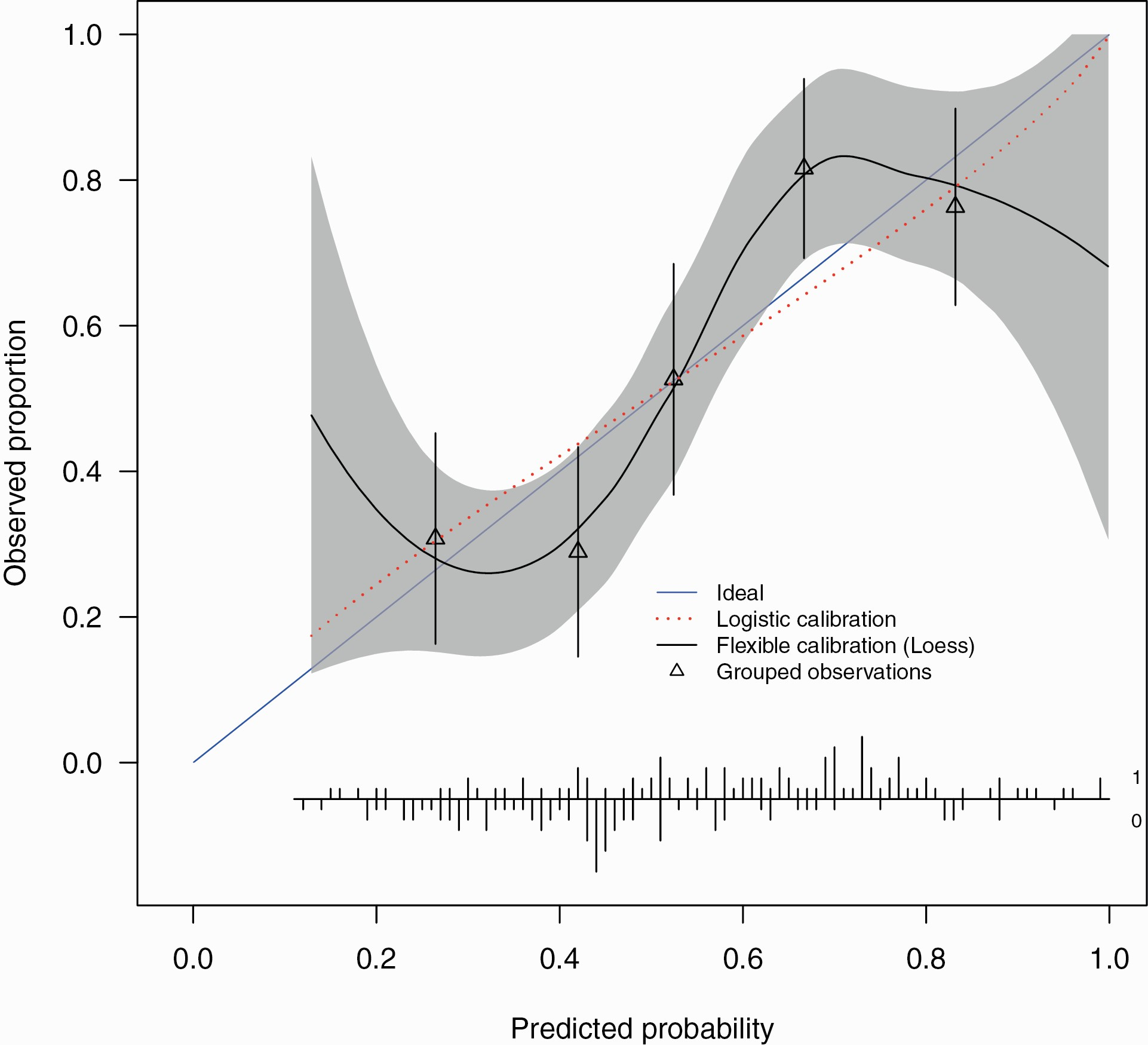}
\caption{Probability calibration plot (reproduced with permission \citep{leighton2021development}) }
\label{fig:ModelPsychosis_CalibrationC}
\end{figure*}

Calibration is typically assessed using flexible calibration curves, which plot estimated risks against observed proportions of events.
Two key measures of calibration are the calibration-in-the-large (alpha) and calibration slope (beta). A well-calibrated model should have an alpha close to zero and a beta close to one. However, these measures alone do not guarantee perfect calibration across all risk levels. Visualization through calibration plots helps identify areas of over- or underestimation. Figure \ref{fig:ModelPsychosis_CalibrationC} shows the calibration curve for the regression model presented in Figure \ref{fig:ModelPsychosis}. This curve provides both qualitative and quantitative assessments of how well the developed risk prediction model aligns with the ideal calibration, represented by a blue line with a slope equal to 1.

Poor calibration can arise from various factors, including differences in patient characteristics or disease prevalence between development and validation populations, changes in healthcare practices over time, model overfitting, and measurement errors in medical data. Strategies to improve calibration include model refitting, continuous updating, and addressing population shifts dynamically. Sample size significantly impacts calibration assessment, with at least 200 events and non-events recommended for precise evaluation. In smaller datasets, evaluating moderate calibration through intercept and slope calculations may suffice.

The importance of calibration in clinical settings cannot be overstated. A poorly calibrated model, even with high discrimination, can lead to misleading or potentially harmful clinical decisions. For example, in the risk prediction of nonremission in first-episode psychosis,overestimation could falsely suggest that patients do not need medication, thereby exposing them to unnecessary risks.

\subsection{Calibration in Deep Learning Models for Clinical Decision Support}

Calibration is also a critical aspect of deep learning models in clinical decision support systems, particularly for establishing trustworthiness with users. A well-calibrated model provides confidence estimates that accurately reflect the probability of correct predictions. For instance, a model with 90\% confidence should be correct 90 out of 100 times.
In practice, perfect calibration is unattainable, but we aim to approximate it. The Expected Calibration Error (ECE) is a common metric used to assess calibration, measuring the difference between confidence and accuracy across prediction bins. 

Recent studies have shown that deeper and more complex neural networks tend to be poorly calibrated, despite high accuracy \citep{guo2017calibration, nixon2019measuring}. Interestingly, increasing model depth or the number of convolutional filters per layer tends to worsen calibration error while improving predictive performance.
For example, a 110-layer ResNet model demonstrated high accuracy but poor calibration, potentially limiting its reliability as a decision support tool. 
The causes of miscalibration in deep networks are not fully understood, but they appear to correlate with model complexity and capacity.
Conversely, techniques like weight decay (L2 regularization) can help reduce calibration error. These findings suggest that in complex networks, over-fitting may manifest in probability estimates rather than classification errors.

For healthcare applications, reliable confidence measures are crucial. Users of clinical decision support systems need to be aware of the confidence level in disease diagnostics. Efforts to improve calibration in deep neural networks include architectural modifications and adjustments to training and optimization strategies. 
While the ECE is widely used, it has limitations. The choice of bin number involves a bias-variance trade-off, and the metric may not fully capture calibration in multi-class problems. Additionally, it can be affected by cancellation effects between over- and under-confident predictions.
Adaptive binning schemes have also been proposed to enhance the stability of calibration measurements.

In conclusion, while deep learning models can achieve high accuracy, their calibration remains a significant challenge. Addressing this issue is essential for developing trustworthy and effective clinical decision support systems. Future research should focus on refining calibration metrics and developing techniques to improve the reliability of confidence estimates in complex neural networks.

\subsection{Decision Curve Analysis}

While accuracy metrics such as sensitivity, specificity, and area under the receiver operating characteristic curve are essential for evaluating prediction models, they fail to capture the clinical consequences of implementing these models in practice. Decision curve analysis (DCA) addresses this limitation by incorporating the concept of net benefit (NB), allowing for a more comprehensive assessment of a model's clinical utility \citep{vickers2006decision, van2018reporting}.

Net benefit is calculated across a range of threshold probabilities (ThresP), representing the point at which a clinician or patient would opt for intervention based on the model's prediction. This approach weighs the benefits of true positive (TP) predictions against the harms of false positives (FP), taking into account the relative importance of these outcomes in a given clinical context.

\begin{equation}
\label{eq:NetBenefit}
NB=\frac{TP}{N} -  \frac{FP}{N} \times \frac{\text{ThresP}}{1 - \text{ThresP}}
\end{equation}

DCA plots the net benefit against threshold probabilities, comparing the prediction model's performance to alternative strategies such as treating all patients or treating none \citep{vickers2019simple}. This visual representation allows stakeholders to assess the model's value across different risk thresholds, which may vary depending on the clinical scenario and individual preferences.

\begin{figure*}
\centering
\includegraphics[width=0.6\linewidth]{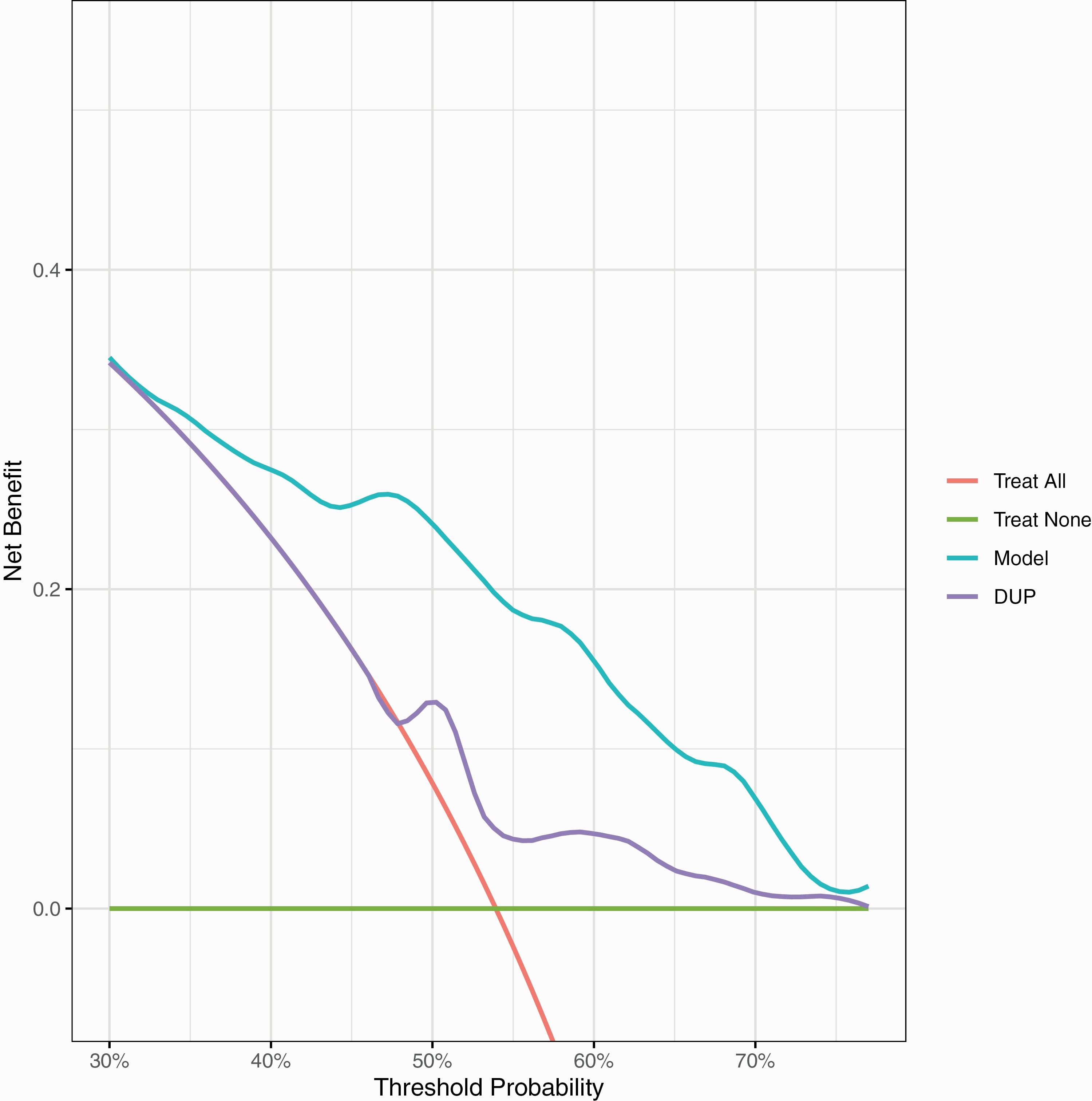}
\caption{Understanding clinical consequences of a risk prediction of non-remission in first-episode psychosis based on Decision Curve Analysis (reproduced with permission \citep{leighton2021development}).}
\label{fig:DecisionCurveAnalysis}
\end{figure*}

For instance, in cancer screening, a lower threshold probability might be preferred due to the severe consequences of missed diagnoses. Conversely, a higher threshold might be appropriate in situations where unnecessary interventions carry significant risks or costs. Figure \ref{fig:DecisionCurveAnalysis} shows the DCA plot for the risk prediction model of nonremission presented in Figure \ref{fig:ModelPsychosis}. The plot demonstrates that the net benefit of using the developed model is higher than the alternatives of treating all, treating no patients, or treating based on the duration of untreated psychosis (DUP). 

The interpretation of DCA results requires careful consideration of the clinical context. A model demonstrating higher net benefit than alternative strategies within a clinically relevant range of threshold probabilities can be considered clinically useful. However, this assessment should be made in conjunction with expert opinion and patient preferences.

DCA can be applied to both continuous probability predictions and binary diagnostic tests. It is particularly valuable when evaluating models in external validation cohorts, as it provides insights into the model's generalizability and potential impact on clinical decision-making.

An illustrative example of DCA in practice is its application to a prediction model for outcomes in first-episode psychosis \citep{leighton2019development, leighton2021development}. By consulting specialist psychiatrists to determine clinically relevant threshold probabilities, researchers were able to demonstrate the model's superior net benefit compared to alternative strategies within the specified range.

It's important to note that while DCA offers valuable insights into clinical utility, it should not replace traditional measures of model performance. Instead, it should be viewed as a complementary tool in the comprehensive evaluation of prediction models, bridging the gap between statistical performance and clinical applicability.

In conclusion, decision curve analysis provides a robust framework for assessing the clinical value of prediction models. By incorporating the concept of net benefit and allowing for comparison across different decision thresholds, DCA enables more informed decisions about model implementation in clinical practice. As healthcare continues to move towards personalized medicine, tools like DCA will play an increasingly crucial role in translating prediction models into meaningful clinical support systems.

\subsection{Responsible Development of Artificial Intelligence-Driven Clinical Decision Support Systems}

Previously, Steyerberg et al. \citep{steyerberg2014towards} has established the importance of four steps to guide the development of machine learning models for healthcare applications: A) - Calibration in the large, B) Calibration slope, C) Discrimination performance established both with internal and external validation and D) Decision-curve analysis. Recently, the integration of artificial intelligence (AI) and in particular deep learning in clinical decision support systems necessitates a more careful consideration of the principles behind responsible model development \citep{de2022guidelines, van2022critical}.  These principles have extended guidelines to include additional safeguards that encompass fairness, explainability, privacy-preservation and interoperability, all of which are crucial for developing trustworthy AI applications in healthcare.

\begin{figure*}
\centering
\includegraphics[scale=0.50]{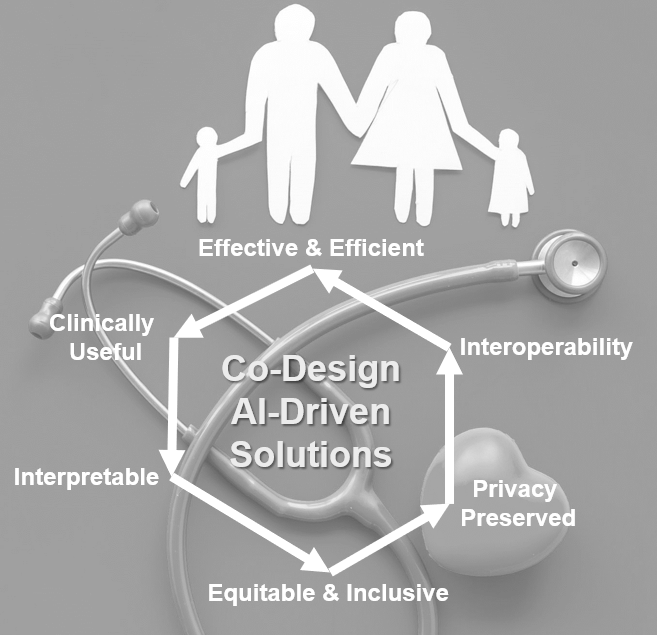}
\caption{Ethical considerations for the use of AI in Clinical Decision Support Systems}
\label{fig:AI_ethics}
\end{figure*}

Figure \ref{fig:AI_ethics} summarises the concepts behind responsible  AI models in healthcare. We already referred to the term 'clinical usefulness', which encapsulates the necessity of an AI model to address a healthcare challenge and its capacity to evaluate if the advantages of an intervention justify the associated risks. ‘Effective and Efficient’ refers to the robustness of the underlying model and whether the outcome is reliably measured as it was highligthed at Steyerberg et al. \citep{steyerberg2014towards}. ‘Interpretability’ is important to understand what the underlying factors are that the model based its specific decision. Interpretability and transparency are cornerstones of responsible AI, enabling users to understand both the technical processes and human decisions involved in the system's operation. Accountability in AI systems requires mechanisms to minimize negative impacts and report adverse consequences. In this context, transparency is crucial for assessing accountability and fairness. 

Healthcare practitioners have expressed concerns about over-reliance on AI systems they cannot fully understand or explain. Addressing these concerns requires further training for practitioners and involving end-users in the design process. The next generation of human-centered clinical decision support systems should possess two key abilities: explaining the model's representation and decisions, and adapting based on feedback.

On the other hand, privacy and data governance are also paramount in clinical decision support systems. Developers must implement safeguards to protect user privacy, ensure data integrity, and control access to sensitive information. 'Privacy-preserving' technologies are also essential to address concerns arising from the use of AI models in new applications that enable real-time tracking of human activity and health in home settings.
Privacy-preserved AI model should leverage interoperable solutions that seamlessly integrate into the healthcare system. 




In conclusion, adherence to responsible AI development guidelines is crucial in designing AI-powered clinical decision support systems. Emphasizing clinical usefulness, explainability, and human-in-the-loop designs is essential to mitigate risks and realize the full potential of AI in healthcare. The development of the right level of interactivity and explainability in medical applications remains an open research question, highlighting the ongoing challenges in this rapidly evolving field.

\section{`Fairness' in Machine Learning Models}
\subsection{Assessing Bias in Clinical Predictive Models}

Electronic health records (EHRs) present unique challenges and opportunities in predictive modeling, particularly in ensuring fairness and equity in healthcare outcomes.
Biases can stem from flaws in study design, execution, or data analysis. To identify such biases, a comprehensive approach is necessary, considering the model's intended use, target population, predictors, and predicted outcomes.

Therefore, the evaluation of predictive models should extend beyond discrimination and calibration to encompass potential biases that may introduce systematic errors. 
A framework for assessing bias in clinical predictive models has been proposed, focusing on four key domains: participant selection, variable and predictor selection, outcome assessment, and analysis \citep{wolff2019probast}. This framework emphasizes the importance of appropriate inclusion and exclusion criteria, consistent predictor definition and assessment across participants, and the use of standardized outcome definitions.

Additionally, the analysis should consider sample size adequacy, handling of continuous and categorical predictors. Researchers are advised to avoid selecting predictors based solely on univariate analysis.
Models can be categorized as having low, medium, or high risk of bias based on the assessment of these domains. Notably, prediction models developed without external validation should generally be considered high risk, except when based on very large datasets.


The concept of "informative presence" in EHRs refers to the potential information carried by the presence or absence of patient data at any given time point. 
For example, EHRs can be inherently biased because sicker individuals are monitored more frequently. This type of informative presence, indicates that the frequency of health records can reflect a patient's health status.
"Informative observation" extends this concept to the timing, frequency, and patterns of longitudinal observations in EHRs, which can provide insights into a patient's evolving health state \citep{sisk2021informative}. While these phenomena can complicate causal or association studies, they also offer potential sources of implicit information that could be exploited in predictive models.

In conclusion, while identifying and addressing bias is crucial for developing robust clinical predictive models, the inherent characteristics of EHRs, such as informative presence and observation, present both challenges and opportunities. Researchers must carefully interpret results while also exploring innovative ways to leverage these implicit data patterns to improve prediction accuracy.

\subsection{Equity Challenges in Machine Learning for Healthcare Applications}

The pursuit of health equity is a global priority, as exemplified by the World Health Organization's vision of a society where all individuals enjoy long, healthy lives \citep{amri2021scoping}. 
Machine learning algorithms have the power to identify unexpected patterns in data, but they can also inadvertently perpetuate and amplify existing biases. This is particularly concerning when these algorithms are used to support clinical decision-making, as they can systematically disadvantage certain population groups \citep{barocas2016big, obermeyer2019dissecting}.

Electronic health records, which serve as the foundation for many predictive models, often reflect historical biases in patient selection, policies, and societal circumstances. These biases can manifest in various ways, such as under-representation of minority groups in the data, systematic differences in feature availability across populations, or the compounding of initial biases over time.

The case of St. George Hospital in the UK serves as a cautionary tale \citep{schwartz2019untold}. In the 1980s, the hospital developed a computer program to streamline medical school admissions based on historical data. Unintentionally, this program formalized existing prejudices against racial minorities and women, demonstrating how algorithmic decision-making can perpetuate systemic biases.

Similar issues have been observed in other domains, such as facial recognition technology and online advertising. These examples highlight the potential for algorithms to discriminate based on race, gender, or age, often due to non-representative training data or the inadvertent use of biased proxies.

In healthcare, a 2019 study revealed that a widely-used risk prediction algorithm considered Black patients to be healthier than equally ill White patients \citep{obermeyer2019dissecting}. This discrepancy arose because the algorithm used healthcare costs as a proxy for health needs, failing to account for disparities in healthcare access and utilization. Consequently, Black patients had to be significantly sicker than White patients to receive the same level of care.

This case underscores the importance of critically examining clinical decision support tools for potential biases. Even when sensitive attributes like race or gender are explicitly excluded from models, they can be implicitly correlated with other features, leading to discriminatory outcomes.

Discriminatory bias in healthcare algorithms can stem from various sources, including study design, data collection, clinician interactions, and patient behaviors. These biases may manifest as label bias, cohort bias, or various forms of data bias, such as minority under-representation or missing data for protected groups.

The interaction between clinicians and predictive models can also introduce biases. Automation bias may lead to over-reliance on model predictions, while dismissal bias could result in ignoring alerts for certain groups. Patient interactions with healthcare systems can further complicate matters, with issues like privilege bias and informed mistrust affecting model effectiveness and fairness.

Addressing algorithmic bias in healthcare is challenging. Simply removing sensitive fields from the data is insufficient, as algorithms can identify and learn from proxy variables. Moreover, the complex nature of these biases makes them difficult to detect and eliminate, particularly when they reflect deeply ingrained societal prejudices.

In conclusion, while machine learning holds great promise for advancing healthcare, it is crucial to remain vigilant about the potential for algorithmic bias. Researchers and practitioners must actively work to quantify and mitigate these biases, ensuring that predictive models promote rather than hinder health equity. This requires ongoing investigation, diverse and representative datasets, and a commitment to fairness in algorithm design and implementation.

\subsection{Strategies to Ensure Fairness in Machine Learning Models for Healthcare}

The pursuit of fairness in machine learning for healthcare applications is crucial, as algorithmic bias can lead to discriminatory outcomes that impact patient care. This section explores systematic approaches to detect and mitigate such biases.

Recent legislation, such as the 2019 Algorithmic Accountability Act in the United States \citep{maccarthy2020examination}, has begun to address these concerns by requiring companies to assess and rectify algorithmic biases. However, the complexity and proprietary nature of many algorithms pose challenges for independent evaluation.

To mitigate discriminatory bias, several strategies have been proposed:
\begin{itemize}
    \item Careful selection and representation of protected groups in the data.
    \item Thorough investigation of potential healthcare disparities in historical data.
    \item Incorporation of fairness goals into model training.
    \item Continuous evaluation of fairness metrics and model performance across groups.
    \item Vigilant monitoring of data and model reassessment during deployment.
\end{itemize}

Furthermore, several fairness metrics have been proposed  \citep{beutel2019putting, majumder2023fair}. Calibration techniques, applied within protected subgroups, can help identify algorithmic bias. Quantifying calibration bias within protected subgroups and assessing statistical parity are essential steps in evaluating model fairness. However, it is crucial to recognize that statistical parity may not be clinically meaningful when disease prevalence differs between subgroups. Other important metrics include independence, separation, and sufficiency \citep{carey2023statistical}. Independence aims for classifier scores to be independent of group membership, while separation focuses on the independence of scores and sensitive variables conditional on the target variable. Sufficiency examines the independence of the target and sensitive variables given a particular score.


It is important to note that these fairness criteria cannot all be satisfied simultaneously when risk prevalence differs across groups. This highlights the need for careful consideration of trade-offs in AI model development.
Therefore, further key objectives for fair decision-making in healthcare are defined that include achieving equal patient outcomes, equal model performance, and equal resource allocation across protected groups. However, these goals may sometimes conflict, necessitating thoughtful prioritization and stakeholder involvement in the design process.

In conclusion, ensuring fairness in machine learning models for healthcare is an ongoing challenge that requires continued research, vigilance, and collaboration among stakeholders. By addressing these issues systematically, we can work towards more equitable and effective healthcare systems that benefit all patients, regardless of their demographic characteristics.


\section{Explainability as a central component of Human-Centered CDSS}

\subsection{Interpretability vs Explainability}

\begin{figure}[h]
    \centering
    \includegraphics[width=0.9\linewidth]{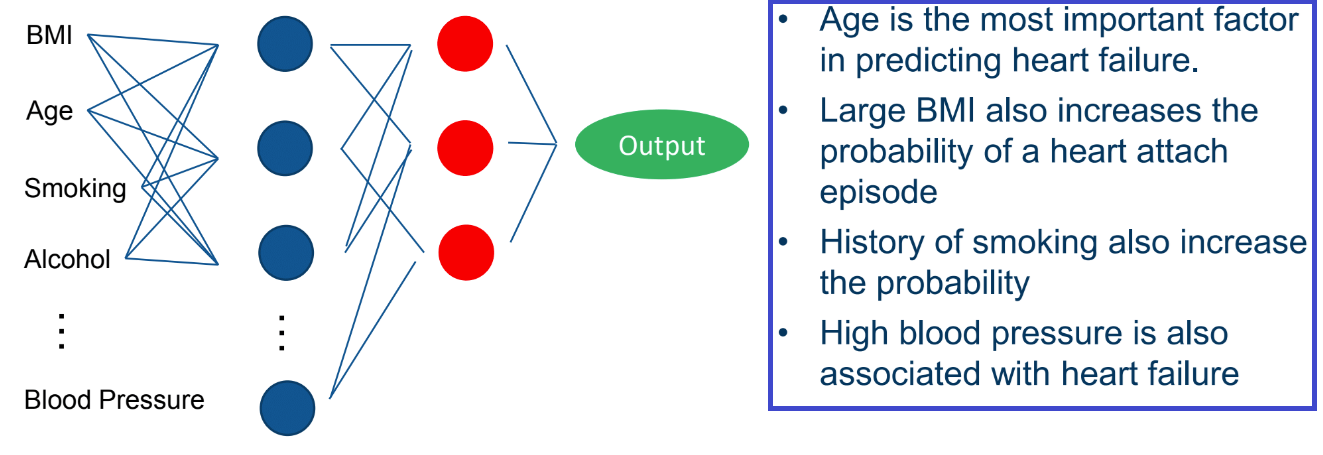}
    \caption{An example network with some specific input factors}
    \label{fig:InterpretModel}
\end{figure}

Interpretability and explainability, while often used interchangeably, have distinct meanings in machine learning. Interpretability is an inherent model property, whereas explainability involves methods to elucidate non-interpretable models. Figure \ref{fig:InterpretModel} illustrates a network assessing patient risk based on factors like BMI, age, smoking habits, alcohol consumption, and blood pressure.   
Consider an 85-year-old female patient with a BMI of 32, high blood pressure, and no smoking or alcohol use. The system labels her 'at risk' recommending medication. However, this output alone may not suffice for a doctor to trust and act upon the model's decision. Understanding the reasoning behind the model's output becomes crucial, highlighting the importance of explainability in complex models, especially in critical domains like healthcare where comprehending the decision-making process is essential for informed and ethical treatment decisions.

Explainability in machine learning models can address crucial questions about a model's performance, success conditions, and decision factors. For example, consider that in heart failure prediction age is a significant factor, with individuals over 60 years having an estimated likelihood of 60\%. Furthermore, a BMI exceeding 25 increases risk by 20\%, as does smoking for over a decade. High blood pressure is also correlated with heart failure. An explainable model should identify these key input factors and quantify their impact on the decision. This insight into the model's underlying function aids result interpretation, clarifies decision-making processes, and helps understand model failures in noisy conditions. Essentially, explainability provides transparency into the model's inner workings, enabling users to comprehend not just what the model predicts, but why it makes those predictions.

\begin{figure}[h]
    \centering
    \includegraphics[width=0.5\linewidth]{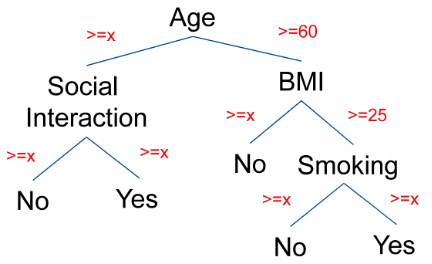}
    \caption{An example of a decision tree}
    \label{fig:DecisionTree}
\end{figure}

Decision trees are widely regarded as highly interpretable machine learning models due to their transparent and intuitive structure as the example at Figure \ref{fig:DecisionTree} shows. The tree-like representation clearly illustrates the decision-making process, with each node representing a specific decision point based on a particular feature. This allows users to easily follow the path from root to leaf, understanding how the model arrives at its predictions. The clear criteria used at each node for splitting the data provide insight into precisely how decisions are made at each step. This hierarchical nature lends itself well to visual representation, making it easier for both experts and non-specialists to grasp the model's logic. For any given prediction, one can trace the exact path through the tree, observing which features were considered and how they influenced the final output. This traceability enhances accountability and facilitates debugging. Moreover, the structure of the tree inherently reveals which features are most important for classification or regression, as they appear closer to the root and in more splits. Unlike more complex models, decision trees can be explained to individuals without a background in machine learning or statistics, facilitating communication between data scientists and stakeholders. 
However, it is important to note that interpretability can decrease as tree depth increases. Very deep trees may become more challenging to interpret, potentially approaching the complexity of "black box" models. 

\begin{figure}[h]
    \centering
    \includegraphics[width=0.9\linewidth]{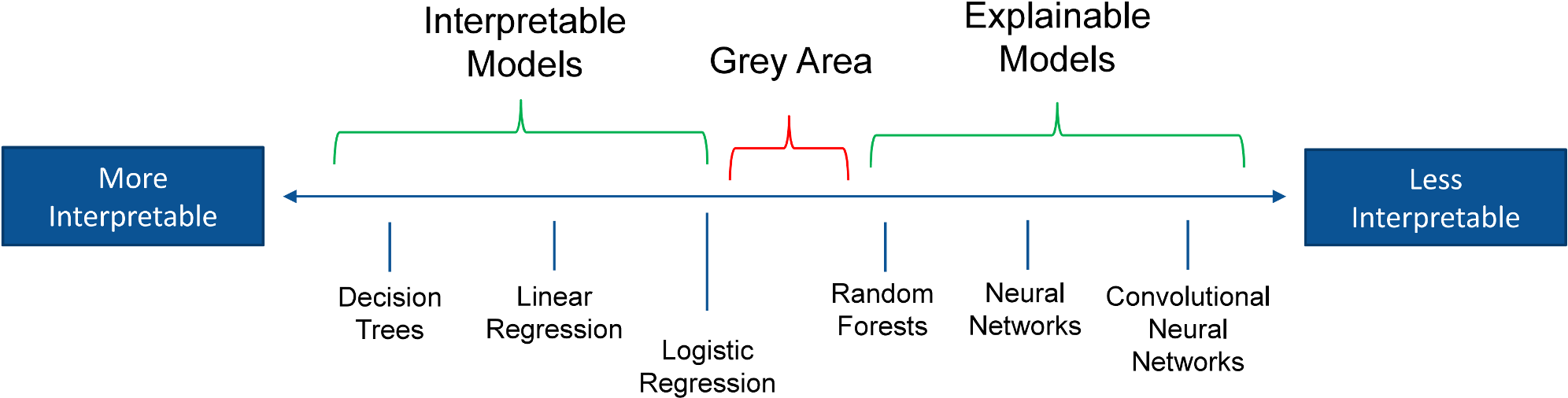}
    \caption{Interpretable vs explainable models}
    \label{fig:InterpretVsExplain}
\end{figure}

Figure \ref{fig:InterpretVsExplain} provides a simplified overview of interpretable and explainable models. On the left, we see inherently interpretable models such as decision trees, linear regression, and logistic regression. These models have been widely used in clinical practice and decision-making due to their simple construction and easily understandable results. The straightforward nature of these models allows practitioners to readily interpret their outputs and understand the reasoning behind decisions. Consequently, these models do not require additional methods to explain their results, as their decision-making process is transparent by design. This intrinsic interpretability distinguishes them from more complex models that may require additional explanation techniques to elucidate their outputs.

While linear regression and decision trees offer high interpretability, they often fall short in performance, especially given the complexity of modern datasets and available computational resources. Sacrificing predictive accuracy for inherent interpretability is increasingly seen as suboptimal in many applications. Instead, the focus has shifted towards developing methods to explain high-performing, complex models. This approach aims to harness the superior predictive power of sophisticated algorithms while still providing insights into their decision-making processes.

\subsection{'Explainability' in Healthcare Applications}


In healthcare applications, explainability is particularly critical for assessing model stability, visualizing relationships affecting outcomes, and enabling ethical analysis, especially concerning minority groups. It also facilitates patient involvement in decision-making processes and allows for the evaluation of privacy risks associated with complex model representations.
This transparency is essential for maintaining the confidence of healthcare professionals, patients, and end-users. Moreover, explainability aids in monitoring model performance over time, as data distributions may shift and affect outcomes.

The significance of explainability extends to various stakeholders. Clinicians require trustworthy models that contribute to scientific knowledge. Patients need assurance of fair treatment and absence of hidden biases. Data scientists and developers utilize explainable models for debugging and improving product efficiency. Management must ensure regulatory compliance, while regulatory bodies need to certify model adherence to legislation.

Explainability is associated with several key objectives, including trustworthiness, causality inference, transferability, informativeness, confidence, fairness, accessibility, interactivity, and privacy awareness. While an explainable model may not guarantee absolute trust or prove causality, it can provide valuable insights into potential causal relationships and help validate results from other inference techniques.

\subsection{Taxonomy of Explainability Methods}

Some of the most common categorisations of explainability methods are local versus global, model agnostic versus model specific, data modality specific versus data modality agnostic and ad-hoc versus post-hoc explanations \citep{arrieta2020explainable, molnar2020interpretable, stiglic2020interpretability}. Local versus global explanation is a very common distinction that translates on whether we get an explanation that relates to the overall function of the model, or an explanation related to specific decisions. Model agnostic versus model specific categorization is also very common \citep{ribeiro2016model}. An important difference is that model agnostic explanations are not bound to a specific machine learning algorithm, whereas model specific explanations are derived directly based on the machine learning model employed. 

Explainability methods can also be categorized based on the modality in the form of data modality specific versus data modality agnostic methods. In this case, the complexity of data representation is highlighted. For example, explaining decisions with relation to imaging data require a completely different approach than explaining differences in tabular data as they do not have any spatio-temporal dependency between variables. Another categorisation is for ad-hoc versus post-hoc explanations. While ad-hoc explanations aim in achieving interpretability by restricting the complexity of the machine learning model (intrinsic explainability), post-hoc methods aim in achieving interpretability by applying methods that analyse the model after training.

\begin{figure}[h]
    \centering
    \subfloat[\centering \label{fig:ModelSpecVsAgnostic}Model specific versus model agnostic explanations] 
    {{\includegraphics[width=0.3\textwidth]{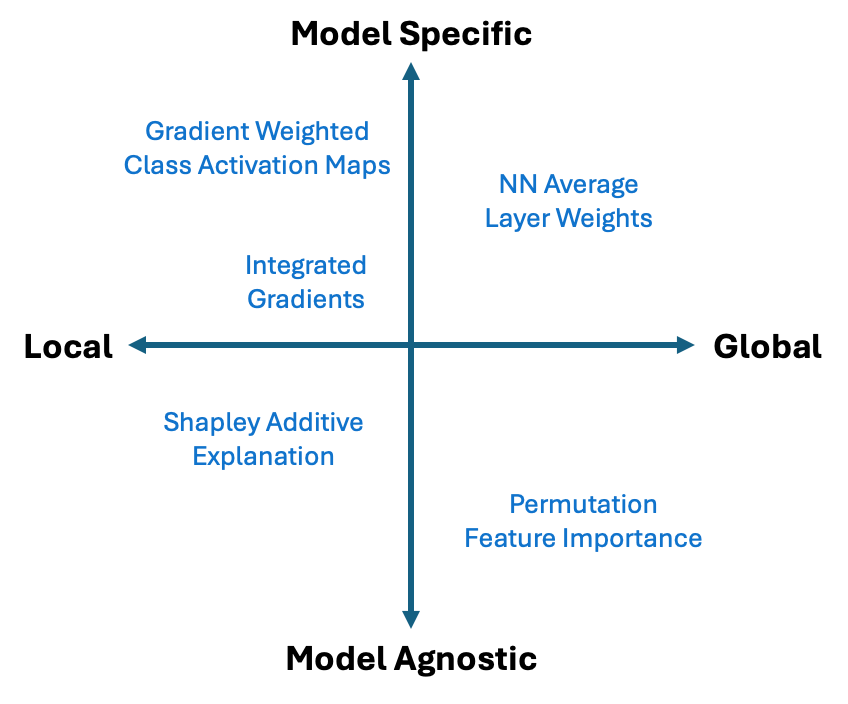} }}
    \hspace{0.1cm}
    \subfloat[\centering \label{fig:OverviewExpl}Overview of the different explainability methods] 
    {{\includegraphics[width=0.6\textwidth]{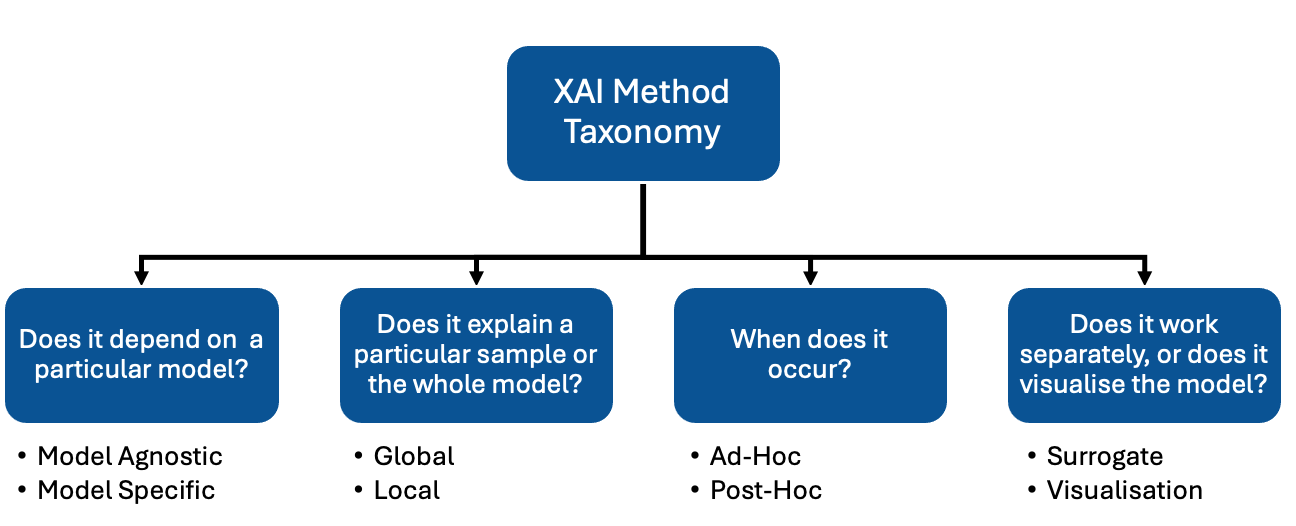} }}
    \caption{Explainability methods.}
    \label{fig:explainabilityMethods}
\end{figure}

Figure \ref{fig:ModelSpecVsAgnostic} presents a two-dimensional taxonomy of explainability methods, organized along two axes: local versus global explanations, and model-specific versus model-agnostic approaches. For example, Shapley Additive Explanation (SHAP) provides both local and global interpretability by analyzing variable impacts through perturbation - locally for individual predictions and globally for overall model behavior. In contrast, class activation maps and integrated gradients provide model-specific interpretability by analyzing neural network activations. These methods trace how specific input features influence the network's decision-making process for individual samples, revealing the internal reasoning patterns of the model.

Permutation feature importance is one of the simplest, yet powerful approaches that links the input variable with the outcome across all the samples and for this reason it is considered a global explainability method.
Permutation based approaches such as permutation feature importance permutes the input values and features in order to understand the importance of each input variable \citep{mi2021permutation}. In other cases, they can apply a surrogate model, an inherently interpretable model that is simple enough to understand its behaviour in a local space. 
The local importance highlights the important features for any individual prediction. It is very helpful to illustrate the local behaviour of the underlying model. Sometimes we can aggregate an average result from local methods in order to get a better idea of how the model functions globally \citep{jones2020improving,MayourianCirculation2024}. 


The advantage that a model-agnostic explanation provides is that we can compare different models with the same data and get an idea of the quality of the prediction as well as the explanation.
Model agnostic tools are usually applied after the model has been trained, so they are considered as post-hoc explanations. In ad-hoc explanations, the model has been designed to be intrinsically explainable. Such an example is representation learning where the model identifies latent factors that intuitively have an explanation and they relate to the function of the model.

On the other hand, we can also categorize methods based on whether they are surrogate models or attribution based methods. Surrogate models basically try to use a simpler model to approximate the behaviour of a black box. Attribution methods can be also categorized based on whether they are perturbation based or back propagation or gradient based methods. 

Figure \ref{fig:OverviewExpl} illustrates the key dimensions for categorizing explainability methods in AI systems. These dimensions include model dependency (agnostic vs. specific), scope of explanation (global vs. local), timing (intrinsic vs. post-hoc), and approach (surrogate models vs. visualization). Each dimension guides the selection of appropriate explainability techniques based on specific analysis requirements.
The diverse taxonomy of explainability approaches yields varied types of model interpretations. Complex AI systems, particularly those serving diverse users, often require multiple complementary explainability methods to provide comprehensive insights into model behavior.

\subsection{Evaluating Explainability in Clinical Decision Support Systems}

The evaluation of explainability methods in clinical decision support systems is a crucial aspect of their development and deployment. This process considers not only the technical aspects of the explanations but also their effectiveness for end-users, whether they are healthcare providers, clinicians, patients or carers.

Explanations typically provide three types of information: the importance of features or attributes to the model, including their interactions; the reasoning behind specific predictions; and an approximation of the complex model using a simpler, interpretable surrogate model such as rule-based systems, decision trees, or linear models.

The evaluation of explanations involves assessing both the model's intrinsic interpretability and the quality of its approximation by interpretable explanations. Key aspects of this assessment include clarity (consistency of rationale for similar instances), parsimony (complexity and compactness of the explanation), fidelity (accuracy in describing the task model), and soundness (truthfulness to the task model).

Attribution-based explanations, which identify the input features most relevant to the model's decision, are common in post hoc explanation methods. While these explanations may not fully satisfy the sufficiency property, they often meet the parsimony criterion if the features are understandable to humans. For clinicians, such explanations are valuable in comparing the model's decision-making process with their own clinical knowledge.

User-based evaluation, both quantitative and qualitative, is crucial in understanding how trust in AI models affects overall system performance in human-in-the-loop scenarios. This approach bridges the gap between technical performance and practical utility in clinical settings, ensuring that explainable AI systems not only perform well mathematically but also integrate effectively into clinical workflows and decision-making processes. Toward this end, quantitative evaluations often utilize metrics based on questionnaires assessing the usefulness, satisfaction, and interest provided by the system's explanations. They may also measure human-machine task performance in terms of accuracy and response time.



\section{Privacy Concerns in Clinical Decision Support Systems: A brief overview}

\subsection{Leakage and privacy concerns of deep learning models}
The integration of machine learning in healthcare applications, particularly in clinical decision support systems (CDSS), necessitates understanding and addressing significant privacy concerns. Advanced machine learning models can memorise and inadvertently expose sensitive information, even when identity data has been removed or pseudo-anonymized. Furthermore, with state-of-the-art sensing technologies, it is now possible to monitor people 24/7 at home for healthcare applications ~\citep{yang2023human}. This capability introduces new privacy challenges, as information about people's activities and physiology can be leaked in real-time ~\citep{zakariyya2024differentially}. For example, wifi and radar-based human motion sensing has gained significant attention for offering unobtrusive observation in sensitive environments like assisted living facilities, hospitals, and residential settings. However, emerging research has uncovered a critical privacy concern: advanced radar systems can now accurately identify individuals through their unique walking patterns. These findings challenge the initial assumption of radar sensing as a privacy-preserving technology, highlighting the urgent need for robust privacy protection mechanisms in sensing systems, even when the raw data may not be visually comprehensible to humans. 
This section explores the inherent risks in deep learning models and discusses strategies to mitigate privacy threats in the context of healthcare applications.

Deep learning models excel at processing complex, correlated sensing data, which has led to substantial improvements in fields such as computer vision and information retrieval. 
However, these models inadvertently memorize training data within their weights, making it possible to reconstruct parts of the original dataset from the algorithm itself ~\citep{hartley2023neural}.  
Traditional anonymization techniques, such as removing personally identifiable information or using pseudo-anonymization, are insufficient protection against sophisticated privacy attacks against deep neural networks. These powerful algorithms can determine an individual's identity by exploiting similarities to other datasets or inferring information from remaining data points. This capability has led to large-scale re-identification attacks.

Privacy threats in machine learning can manifest in various sophisticated ways, even with limited model access. For example, memorisation of the training data might be reflected in assigning high likelihood to specific input samples.
An attacker might gain insights into a model through its architecture, weights, training data, or even just its output logits. In the context of membership inference attacks (MIA), adversaries can potentially determine whether a specific data record was part of the original training dataset. The attack typically involves an attacker model that learns to predict the outcomes of the target model, with the goal of reconstructing sensitive input data. 

While seemingly straightforward, MIA attacks serve as a foundational technique for more advanced data extraction methods via model inversion and reconstruction. 
The attack process often employs a shadow model, which is a surrogate model with similar architecture as the target model but different parameters ~\citep{shokri2017membership}. By generating synthetic data and evaluating its performance against the shadow model, the attacker can identify data likely similar to the original training set. This method is particularly powerful when the attacker lacks direct access to the target model's training data. A high prediction score from the shadow model suggests that the synthetic data closely resembles the original target data ~\citep{carlini2022membership}.



To address these concerns, innovative approaches are being developed. One promising method involves disentangling latent representations in data, separating key features from identity information ~\citep{Podjaski2021,li2024mutual}. This approach has shown success in improving classification performance while protecting individual privacy, as demonstrated in recent work on human pose data. Other potential mitigation mechanisms include incorporating robust privacy techniques during the development of the target model to enhance its resistance to data leakage attacks.

In summary, the powerful memorization capabilities of deep neural networks create inherent vulnerabilities that can compromise patient privacy. It is essential to safeguard user privacy by integrating robust privacy protection mechanisms early in the development of decision support systems. Tools like privacy-meter available at ~\citep{murakonda2020ml} can facilitate this process by measuring how an AI system may potentially leak sensitive information.  By separating biometric data from features of interest and filtering out identity information early in the processing pipeline, it is possible to enhance both the effectiveness and the privacy-preserving qualities of these systems.


\subsection{ Defenses against privacy attacks}
Privacy preservation in healthcare machine learning applications is crucial, both at the data and model levels. This section explores various approaches to protect patient privacy, contrasting centralized and federated learning methods.

\subsubsection{Differential Privacy Against privacy attacks}

\textbf{Differential Privacy (DP)} is a mathematical framework designed to protect sensitive data used in training AI models. DP algorithms mitigate the risk of data leakage by introducing calibrated noise into computations. This is particularly valuable when handling sensitive information, such as healthcare data, in the development of decision support systems. The procedure involves adding noise based on two privacy budget parameters, $\epsilon$ and $\delta$ ~\citep{dwork2014algorithmic}. The parameter $\epsilon$ controls the amount of noise added, while $\delta$ defines the probability of the mechanism failing to maintain privacy. Stronger privacy guarantees are achieved with smaller values of  $\epsilon$ and $\delta$. \\

\noindent\textbf{Definition 1}\label{Def_1}: \textit{A randomized function $F$ provides pure $\epsilon$-DP
if for all neighbouring input datasets $D_1$ and $D_2$ differing on at most one element and $\forall S  \subseteq Range(F)$} ~\citep{dwork2014algorithmic}, satisfying equation \ref{eqdp1} below, where $\Pr$ in equations \ref{eqdp1} and \ref{eqdp2} represents a probability measure. 

\begin{equation}
\label{eqdp1}
\Pr[F(D_1) \in S] \leq e^\epsilon \Pr[F(D_2) \in S]
\end{equation}

For ($\epsilon, \delta$)-DP, the guarantee is relaxed as follows:

\begin{equation}
\label{eqdp2}
\Pr[F(D_1) \in S] \leq e^\epsilon \Pr[F(D_2) \in S] + \delta
\end{equation}

Thus, when $\delta = 0$, the relaxed guarantee simplifies to the pure $\epsilon$-DP condition. The noise introduced can follow either a Gaussian or Laplace distribution, depending on the desired balance between privacy guarantees and utility. DP techniques are extensively utilized to develop robust machine learning models that protect training data from leakage, especially in the presence of MIA. 
In this context, noise can be introduced to either the data or the model gradients during training.

DP methods that focus on input data features often rely on the addition of Laplace noise ~\citep{phan2017adaptive, fujimoto2023differential, zakariyya2024differentially}. Proper implementation of additive noise injection on specific features effectively safeguards sensitive data against MIA ~\citep{zakariyya2024differentially}.

Another widely adopted mechanism for creating DP-compliant models is applying $(\epsilon, \delta)$-DP to model gradients during training ~\citep{abadi2016deep, dupuy2022efficient, boenisch2024have}. This approach involves clipping gradients and adding Gaussian noise proportional to $\epsilon$ at each training iteration. The clipping norm, a fixed threshold (C), bounds the gradient magnitudes for individual data points, managing the effect of large gradients ~\citep{kong2024unified}. The model is iteratively trained using mini-batches of the input data samples and a stochastic gradient descent (SGD) algorithm, commonly referred to as the DP-SGD procedure. Algorithm \ref{alg:dp_sgd} outlines the DP-SGD training procedure, incorporating gradient clipping and noise addition to ensure DP guarantee.

\begin{algorithm}[H]
\caption{Differentially Private Stochastic Gradient Descent (DP-SGD)}
\label{alg:dp_sgd}
\textbf{Input:} Dataset $\mathcal{D} = \{x_1, \dots, x_n\}$, loss function $\mathcal{L}$, learning rate $\eta$, clipping norm $C$, noise scale $\sigma$, batch size $B$, number of iterations $T$.\\
\textbf{Output:} Model parameters $\theta_T$.
\begin{algorithmic}[1]
\STATE Initialize model parameters $\theta_0$.
\FOR{$t = 1$ to $T$}
    \STATE Sample a random mini-batch $\mathcal{B}_t \subset \mathcal{D}$ of size $B$.
    \FOR{each $x_i \in \mathcal{B}_t$}
        \STATE Compute the gradient $\mathbf{g}_i = \nabla_\theta \mathcal{L}(\theta_{t-1}, x_i)$.
        \STATE Clip the gradient: $\mathbf{\tilde{g}}_i = \mathbf{g}_i / \max\left(1, \frac{\|\mathbf{g}_i\|_2}{C}\right)$.
    \ENDFOR
    \STATE Aggregate the clipped gradients: $\mathbf{\bar{g}} = \frac{1}{B} \sum_{i \in \mathcal{B}_t} \mathbf{\tilde{g}}_i$.
    \STATE Add noise: $\mathbf{\hat{g}} = \mathbf{\bar{g}} + \mathcal{N}(0, \sigma^2 C^2 \mathbf{I})$.
    \STATE Update the model: $\theta_t = \theta_{t-1} - \eta \mathbf{\hat{g}}$.
\ENDFOR
\RETURN $\theta_T$
\end{algorithmic}
\end{algorithm}

\subsubsection{Federated Learning and Defenses Against Privacy Attacks}


Traditionally, machine learning models in healthcare have been developed in centralized settings, where both data and models reside within the same environment ~\citep{alanazi2022using}. For instance, the MIMIC database, a clinical database integrating information from thousands of patients exemplifies such centralized approaches. Similarly, technology corporations like Google collect usage data directly from mobile devices for training models. While these methods are effective in leveraging large datasets, they also raise significant privacy concerns.

Federated learning has emerged as an alternative decentralized approach for building machine learning models. This paradigm involves a single server and multiple clients, each retaining their own data. Instead of centralizing data, the algorithm is distributed to where the data resides. Training iterations occur locally on the clients, with model parameters shared with a centralized server that aggregates these updates to create a global model ~\citep{mcmahan2017communication}. 
A commonly used federated learning algorithm is Federated Averaging (FedAvg) ~\citep{mcmahan2017communication}. FedAvg enables the creation of a global model by using a central server to average the weights of local models from various client devices. This process is iterative: during each communication round, the server interacts with clients to update the global model until convergence is achieved. 

Federated learning allows data to remain with its owners while still enabling collaborative algorithm training. However, it can be communication and memory intensive. Additionally, federated learning alone does not inherently guarantee security and privacy, necessitating additional protective measures. Both centralized and edge models are vulnerable to privacy and adversarial attacks ~\citep{Kaissis2020,song2020analyzing, kumar2023impact}.



Model-level defenses include machine unlearning (or "forgetting"), which aims to remove specific data without retraining the entire model. This approach, while conceptually appealing, faces challenges in implementation and verification. Adversarial defenses, such as adding targeted noise to confidence score vectors, have been proposed to protect against MIA, though they lack theoretical privacy guarantees ~\citep{jia2019memguard}.

DP has gained significant attention in federated learning due to its ability to provide a guaranteed maximum privacy loss. In federated learning, this method involves adding calibrated noise to model updates during local training on each device, enhancing both generalization and patient privacy. However, determining appropriate privacy thresholds for local model updates remains a challenge.

Homomorphic encryption represents another promising avenue for secure AI in healthcare, allowing data processing without decryption \citep{Kaissis2020}. However, its computational complexity is prohibitive for practical applications  and it also suffers from a lack of explainability and transparency. Despite these challenges, it offers strong privacy guarantees and it is considered a key element in next-generation secure and private AI systems. 



Ultimately, the goal is to strike an optimal balance between accuracy, explainability, fairness, and privacy in healthcare AI systems. While federated learning and other privacy-preserving methods offer promising solutions, they must be carefully combined and implemented to ensure both the utility and security of sensitive healthcare data.
Several open questions remain in this field. Researchers are exploring whether decentralized data storage and federated learning can enable cross-institutional research while preserving privacy.

\subsection{Adversarial attacks against explanations in Deep Learning}


Explainability techniques in deep neural networks, aimed at improving interpretability and trust, are increasingly recognized as being susceptible to adversarial attacks ~\citep{baniecki2024adversarial}. This section examines the vulnerabilities of these techniques and their potential implications for healthcare applications, where trust and reliability are paramount.


Recent research has revealed that explanations for deep learning models can be manipulated in ways that are difficult to detect. Two main types of attacks have been identified: those that manipulate the model's loss function to produce misleading explanations while preserving performance, and those that introduce small, nearly imperceptible perturbations to input data to alter explanations significantly. Popular adversarial attacks targeting explainability models include both white-box and black-box approaches ~\citep{baniecki2024adversarial, vadillo2024adversarial}. White-box attacks operate under the assumption that the adversary has complete knowledge of the model's architecture, parameters, and gradients. Examples of white-box attacks include the Fast Gradient Sign Method (FGSM) ~\citep{goodfellow2014explaining}, its iterative variant, the Projected Gradient Descent (PGD) ~\citep{kurakin2016adversarial}, and the Carlini \& Wagner (C\&W) attack ~\citep{carlini2017towards}. These methods leverage detailed internal information to craft adversarial examples that maximize model vulnerability.

In contrast, black-box attacks are designed for scenarios where the attacker lacks access to the model's internal parameters or architecture. These attacks are often query-based, relying on input-output observations to infer model behavior. A notable example is the Zero Order Optimization (ZOO) attack ~\citep{chen2017zoo}, which uses iterative querying to approximate gradients and generate adversarial examples.

Both attack paradigms aim to manipulate model gradients, creating adversarial inputs that lead to misclassifications. Such attacks pose significant risks to the security and robustness of AI systems, particularly in sensitive applications where model integrity is important.

Gradient-based explanation methods, such as Grad-CAM, are particularly susceptible to manipulation. Researchers have demonstrated various techniques to fool these methods, including:
\begin{itemize}
\item Location fooling: Making the explanation highlight a specific region of the input, regardless of its relevance.
\item Top-k fooling: Reducing the importance of pixels that originally had the highest interpretation scores.
\item Center-mass fooling: Optimizing the heatmap to diverge as much as possible from the original without affecting classification performance.
\item Active fooling: Swapping explanations between different target classes entirely.
\end{itemize}
These manipulations are achieved by introducing additional terms to the loss function during training, balancing between classification accuracy and explanation manipulation.

Research has identified potential defenses. One approach involves smoothing explanations at the network level by replacing rectified linear unit (ReLU) activation functions with softplus functions. The smoothness of these functions, controlled by a beta parameter, can be adjusted to make networks more resistant to data perturbation-based manipulations. DP can also help protect against privacy attacks on explanations.

Understanding and modeling the susceptibility of neural networks to explanation manipulation is crucial. It not only helps in preventing malicious attacks but also in detecting when explanations might be misleading due to other limitations. This knowledge can guide the development of more robust explainability techniques, which is particularly important in healthcare applications where the reliability of AI-assisted decisions is paramount.


\subsection{Trade-offs Between Privacy Protection and Model Performance}

The implementation of privacy-preserving techniques in machine learning introduces a fundamental tension between data protection and algorithmic performance. This delicate balance is particularly critical in healthcare applications, where both privacy and predictive accuracy are paramount.
Privacy-preserving methods fundamentally challenge the traditional approach to machine learning by introducing mechanisms that deliberately obscure and modify data or alter the convergence of the model training process. At the core of this challenge lies the complex interplay between protecting individual privacy and maintaining the integrity of predictive models. When privacy techniques are applied, they invariably reduce the richness and depth of available information, creating a nuanced landscape of compromises.

Feature representation suffers particularly from privacy sanitization techniques. By removing or masking sensitive identifying information, models lose critical contextual information that might be essential for accurate predictions. In personalized medicine, this can mean overlooking subtle but important correlations that could significantly impact diagnostic accuracy.

Empirical research has demonstrated concrete performance implications. Differential privacy can reduce model accuracy by 5 to 20 percent, depending on the privacy budget. They are also notoriously difficult to train from scratch and thus they depend on pretrained models.  Federated learning models often show a 3 to 15 percent performance reduction compared to centralized training approaches. Synthetic data generation presents even more significant challenges, potentially leading to up to 30 percent loss in predictive power if not meticulously implemented.

Despite these technical challenges, the ethical imperative remains clear. The potential psychological and social harm from data breaches far outweighs marginal improvements in predictive accuracy. Privacy protection must remain a fundamental consideration, not an afterthought. Ultimately, the goal is not to choose between privacy and performance, but to develop intelligent systems that can maintain both. 

\section{Conclusions}

The integration of artificial intelligence into clinical decision support systems represents both a transformative opportunity and a complex challenge in healthcare. While these systems offer unprecedented capabilities for improving patient care, their successful implementation demands a careful balance of multiple critical factors. The development pathway must address not only technical excellence in model performance, but also robust validation, proper calibration, and thorough decision curve analysis. As we look to the future, the success of AI in healthcare will depend on our ability to navigate challenges in explainability, causality, and bias while maintaining the trust of both healthcare providers and patients.

The complexity of healthcare AI systems raises three paramount concerns that must be carefully balanced. First, there is the critical challenge of ensuring fairness and addressing bias, particularly given the inherent "informative presence" in electronic health records and their potential to reflect historical societal disparities. Second is the need for transparency and explainability, requiring sophisticated approaches that go beyond simple interpretability to provide meaningful insights for healthcare professionals and patients alike. Third is the fundamental requirement to protect patient privacy, particularly given the potential vulnerabilities created by deep learning models' ability to memorize training data. While techniques such as differential privacy, federated learning, and homomorphic encryption offer promising solutions, they often require careful trade-offs between privacy protection and model performance.

Success in this evolving landscape requires a holistic approach that considers these interconnected challenges. The path forward demands ongoing collaboration between technical experts, healthcare providers, and policymakers to develop systems that are not only technically sophisticated but also ethically sound, clinically useful, and privacy-preserving. As AI continues to transform healthcare, our ability to balance these competing demands while maintaining focus on improved patient outcomes will determine the ultimate impact of these innovations on the future of medicine.\\

\noindent{\textbf{Acknowledgements}}
Fani Deligianni is supported by funding from EPSRC (EP/W01212X/1) and Academy of Medical Sciences (NGR1/1678). She is also a member of the research team for NIHR (NIHR158303). This chapter builds upon content from the Coursera specialization 'Informed Clinical Decision Making using Deep Learning'.

\newpage
\bibliographystyle{unsrtnat}
\bibliography{references}

\end{document}